\documentclass{article}

\usepackage{aaai}
\usepackage{times}
\usepackage{helvet}
\usepackage{courier}
\usepackage{amssymb,amsmath}
\usepackage{a4,a4wide}

\usepackage[pdftex]{graphicx}

\usepackage{url}

\newcommand{\lrarrow}{\leftrightarrow}
\newcommand{\lrarrows}{\leftrightarrow_s}
\newcommand{\la}{\leftarrow}

\newcommand{\F}{\ensuremath{Forget}}
\newcommand{\IR}{{\ensuremath {I\!R}}}

\newcommand{\alphabet}{\ensuremath{\mathcal{A}}}
\newcommand{\naf}[1]{\ensuremath{{\sim}{#1}}}
\newcommand{\head}[1]{\ensuremath{H(#1)}}

\newcommand{\body}[1]{\ensuremath{\mathit{B}(#1)}}
\newcommand{\poslits}[1]{\ensuremath{#1^+}}
\newcommand{\neglits}[1]{\ensuremath{#1^-}}

\newcommand{\AS}[1]{\ensuremath{\mathit{AS}(#1)}}
\newcommand{\SE}[1]{\ensuremath{\mathit{SE}(#1)}}
\newcommand{\SEs}[2]{\ensuremath{\mathit{SE}_{#1}({#2})}}
\newcommand{\equivs}{\equiv_s}
\newcommand{\incls}{\models_s}
\newcommand{\vdashs}{\vdash_s}

\newcommand{\Sig}[1]{\sigma(#1)}
\newcommand{\K}{{\mathcal K}}

\newcommand{\Mod}[1]{{\mathit{Mod}({#1})}}

\newcommand{\Cns}[2]{\mbox{${\cal C}n_{#1}(#2)$}}

\newcommand{\calS}{\ensuremath{\mathcal{S}}}
\newcommand{\forget}[2]{\ensuremath{\mathsf{forget}(#1,#2)}} %
\newcommand{\dru}{\ensuremath{\mathit{.\quad}}}

\newcommand{\reduce}[2]{\ensuremath{#1_{\mid #2}}}
\newcommand{\expand}[2]{\ensuremath{#1_{\uparrow #2}}}

\newtheorem{theorem}{Theorem}
\newtheorem{definition}{Definition}
\newtheorem{proposition}{Proposition}

\newtheorem{example}{Example}

\title{An Approach to Forgetting in Disjunctive Logic Programs that
Preserves Strong Equivalence}

\author{James P.\ Delgrande  \\
School of Computing Science \\
Simon Fraser University  \\
Burnaby, B.C. V5A 1S6\\
Canada \\
jim@cs.sfu.ca
\And
Kewen Wang\\
School of Information and Communication Technology\\
Griffith University,\\
Brisbane, QLD 4111 \\
Australia \\
k.wang@griffith.edu.au
}

\begin{document}
\nocopyright
\maketitle

\begin{abstract}
In this paper we investigate forgetting in disjunctive logic programs,
where forgetting an atom from a program amounts to a reduction in the
signature of that program.
The goal is to provide an approach that is syntax-independent,
in that if two programs are strongly equivalent, then the results of
forgetting an atom in each program should also be strongly equivalent.
Our central definition of forgetting is impractical but satisfies this
goal:
Forgetting an atom is characterised by the set of SE consequences of the
program that do not mention the atom to be forgotten.
We then provide an equivalent, practical definition, wherein forgetting an
atom $p$ is given by those rules in the program that don't mention $p$,
together with rules obtained by a single inference step from rules that do
mention $p$.
Forgetting is shown to have appropriate properties;
as well, the finite characterisation results in a modest (at worst
quadratic) blowup.
Finally we have also obtained a prototype implementation of this approach
to forgetting.  
\end{abstract}

\section{Introduction}
\label{sec:introduction}

\emph{Forgetting} is an operation for eliminating variables from a knowledge
base \cite{LinReiter94,LLM03}. 
It constitutes a reduction in an agent's language or, more accurately,
signature, and has been studied under different names, such as variable
elimination, uniform interpolation and relevance \cite{GreinerEtAl97}.
Forgetting has various potential uses in a reasoning system.
For example, in query answering, if one can determine what is relevant to a
query, then forgetting the irrelevant part of a knowledge base may yield a
more efficient operation.
Forgetting may also provide a formal account and justification of
predicate hiding, for example for privacy issues.
As well, forgetting may be useful in summarising a knowledge base or reusing
part of a knowledge base or in clarifying relations between predicates.

The best-known definition of forgetting is with respect to classical
propositional logic, and is due to George Boole \cite{Boole1854}. 
To forget an atom $p$ from a formula $\phi$ in propositional logic, one
disjoins the result of
uniformly substituting $\top$ for $p$ in $\phi$ with the result of
substituting $\bot$; that is, forgetting is given by $\phi[p/\top] \vee  \phi[p/\bot]$.
\cite{LinReiter94} investigated the theory of forgetting for first order logic and its application in reasoning about action.
Forgetting has been applied in resolving conflicts \cite{EiterWang08,zhafoo97a},
and
ontology comparison and reuse \cite{KontchakovWZ08,KontchakovWZ10}. 

The knowledge base of an agent may be represented in a non-classical logic,
in particular a nonmonotonic approach such as answer set programming (ASP)
\cite{GelfondLifschitz88,Baral03,GebserEtAl12}.
However, the Boole definition clearly does not extend readily to logic programs.
In the past few years, several approaches have been proposed for forgetting in ASP \cite{EiterWang06,EiterWang08,WangSattarSu05,ZhangFW05,ZhangFoo06}. 
The approach to forgetting in \cite{ZhangFW05,ZhangFoo06} is syntactic,
in the sense that their definition of forgetting is given in terms of
program transformations, but is not based on answer set semantics or
SE models%
\footnote{See the next section for definitions.}
(for normal logic programs). 
A semantic theory of forgetting for normal logic programs under answer set
semantics is introduced in \cite{WangSattarSu05}, in which a sound and
complete algorithm is developed based a series of program transformations.
This theory is further developed and extended to disjunctive logic
programs \cite{EiterWang06,EiterWang08}.
However, this theory of forgetting is defined in terms of standard answer set
semantics instead of SE models.

In order to use forgetting in its full generality, for dealing with
relevance or predicate hiding, or in composing, decomposing, and reusing
answer set programs, it is desirable for a definition to be given in terms of
the \emph{logical content} of a program, that is in terms of SE models.
For example, the reuse of knowledge bases requires that
when a sub-program $Q$ in a large program $P$ is substituted with another program $Q'$, the resulting program should be equivalent to $P$. This is not the case for answer set semantics due to its nonmonotonicity.   
As a result, two definitions of forgetting have been introduced in HT-logic \cite{WangZZZ12,WangWangZhang13}.
These approaches indirectly establish theories of forgetting under SE models as HT-logic provides a natural extension of SE models. The approach to interpolation for equilibrium logic introduced in \cite{GabbayPV11} is more general than forgetting.
However, the issue of directly establishing a theory of forgetting for disjunctive logic programs under SE models is still not fully resolved yet.
In addition, it is even more challenging to develop efficient algorithm for computing a result of forgetting under SE models.

A key intuition behind forgetting is that the logical consequences of a set of formulas that don't mention
forgotten symbols should still be believed after forgetting.
This leads to a very simple (abstract) knowledge-level definition, provided that a consequence operator is provided in the underlying logic.
In particular, the semantics of a logic usually associates a set of models $\Mod{\K}$ with each knowledge base $\K$.
This makes it straightforward to formulate a definition of forgetting based on the above intuition. 
However, such a definition of forgetting suffers from the problem of inexpressibility, i.e., the result of forgetting may not be expressible in the logic.
In this paper, we establish such a theory of forgetting for disjunctive logic programs under SE models.
Besides several important properties, we show that the result of forgetting for a given disjunctive program is still a disjunctive program.
This result confirms the existence and expressibility of forgetting for DLP
under SE models and in fact provides an algorithm for computing forgetting under SE models.
We investigate some optimisation techniques for the algorithm and   
report a prototype implementation of the algorithm.

\section{Answer Set Programming}

Here we briefly review pertinent concepts in answer set programming;
for details see \cite{GelfondLifschitz88,Baral03,GebserEtAl12}.

Let $\alphabet$ be an alphabet, consisting of a set of \emph{atoms}.
A (\emph{disjunctive}) \emph{logic program} over $\alphabet$ is a finite
set of {rules} of the form
\begin{equation}\label{eq:r}
a_1;\dots;a_m
\la
b_1,\dots,{b_n},
\naf{c_1},\cdots,\naf{c_p}.
\end{equation}
where $a_i,b_j,c_k\in\alphabet$, and $m, n, p \geq 0$ and $m+n+p > 0$.
Binary operators `;' and `,' express disjunction and conjunction respectively.
For atom $a$, $\naf a$ is (default) negation.
%representing negation as failure.
We will use ${\cal L}_\alphabet$ to denote the language (viz. set of rules)
generated by $\alphabet$.

Without loss of generality, we assume that there are no repeated literals in a rule.
The \emph{head} and \emph{body} of a rule $r$, $\head{r}$ and $\body{r}$, are
defined by:
\begin{eqnarray*}
\head{r} &=& \{a_1,\dots,a_m\} \qquad \mbox{and}
\\
\body{r} &=& \{{b_1},\dots,{b_n},\naf{c_1},\dots,\naf{c_p}\}.
\end{eqnarray*}
Given a set $X$ of literals, we define
\begin{eqnarray*}
\poslits{X} &=& \{a\in\mathcal{A}\mid a\in X\},
\\
\neglits{X} &=& \{a\in\mathcal{A}\mid \naf{a}\in X\},\mbox{ and}
\\
\naf{X} &=& \{\naf{a} \mid a\in X\cap\mathcal{A}\}.
\end{eqnarray*}
For simplicity, we sometimes use a set-based notation, expressing a rule
as in~(\ref{eq:r}) as
\[
H(r)\!\leftarrow\poslits{B(r)},\naf{\neglits{B(r)}}
\ .
\]
The \emph{reduct} of a program~$P$ with respect to a set of atoms $Y$,
denoted $P^Y$, is the set of rules:
%\[
%\{
%\poslits{H(r)}\!\leftarrow\poslits{B(r)}
%\mid 
%r\in P, \; \neglits{H(r)}\subseteq Y, \; \neglits{B(r)}\cap Y=\emptyset
%\}.
%\]
\[
\{
H(r)\!\leftarrow\poslits{B(r)}
\mid 
r\in P,\; \neglits{B(r)}\cap Y=\emptyset
\}.
\]
Note that the reduct consists of negation-free rules only.
An \emph{answer set} $Y$ of a program~$P$ is a subset-minimal model of $P^Y$.
A program induces 0, 1, or more \emph{answer sets}.
The set of all answer sets of a program $P$ is denoted by \AS{P}.
For example, the program
\(
P=\{a\la.\quad c;d\la a,\naf{b}\}
\)
has answer sets $\AS{P}=\{\{a,c\},\{a,d\}\}$.  
Notably, a program is nonmonotonic with respect to its answer sets.
For example, the program
$\{q \leftarrow \naf p\}$ has answer set $\{q\}$ while
$\{q \leftarrow \naf p. \;\; p \leftarrow\}$ has answer set $\{p\}$.

\subsection{SE Models}
As defined by \cite{Turner03},
an \emph{SE interpretation} on a signature $\alphabet$ is a pair $(X,Y)$ of interpretations such that $X\subseteq Y\subseteq\alphabet$.
An SE interpretation is an \emph{SE model} of a program $P$ if $Y\models P$ and
$X\models P^Y$, where $\models$ is the relation of logical entailment in
classical logic.
The set of all SE models of a program $P$ is denoted by $\SE{P}$.
%this notation extends in the obvious fashion to the set of SE models for a
%given alphabet, viz.\ $\SE{\alphabet}$.
%
Then, $Y$ is an answer set of $P$ iff $(Y,Y)\in\SE{P}$ and no
$(X,Y)\in\SE{P}$ with $X\subset Y$ exists.
Also, we have $(Y,Y)\in\SE{P}$ iff $Y\in\Mod{P}$.

A program $P$ is \emph{satisfiable} just if $\SE{P} \neq \emptyset$.%
\footnote{Note that many authors in the literature define satisfiability in terms of
answer sets, in that for them a program is satisfiable if it has an answer
set, i.e., $\AS{P} \neq \emptyset$.}
Thus, for example, we consider $P = \{ p \la \naf{p}\}$ to be satisfiable,
since  $\SE{P} \neq \emptyset$ even though $\AS{P} = \emptyset$.
Two programs $P$ and $Q$ are \emph{strongly equivalent}, symbolically
$P \equivs Q$, iff $\SE{P}=\SE{Q}$.
Alternatively, $P \equivs Q$ holds iff $\AS{P\cup R}=\AS{Q\cup R}$, for
every program $R$ \cite{LifschitzPearceValverde01}.
We also write $P\incls Q$ iff $\SE{P}\subseteq\SE{Q}$.
%For simplicity, we often drop set-notation within SE interpretations
%and simply write, e.g., $(a,ab)$ instead of $(\{a\},\{a,b\})$.

\subsection{SE Consequence}

While the notion of SE models puts ASP on a monotonic footing with respect
to model theory, \cite{Wong2008} has subsequently provided an inferential
system for rules that preserves strong equivalence, where his notion of
\emph{SE consequence} is shown to be sound and complete with respect to the
semantic notion of \emph{SE models}.
His inference system is given as follows, where lower case letters are
atoms, upper case are sets of atoms, and for a set of atoms
$C = \{ c_1, \dots, c_n\}$, $\naf C$ stands for
$\{ \naf c_1, \dots, \naf c_n\}$.

\medskip

\noindent
{\bf Inference Rules for SE Consequence}:
\begin{description}
\item[Taut]
$x \la x$

\item[Contra]
$\la x, \naf x$

\item[Nonmin]
From
$A \la B, \naf C$
\quad
infer

\qquad\qquad
$A ; X \la B, Y, \naf C, \naf Z$

\item[WGPPE]
From
\mbox{$A_1 \!\la\! B_1, x, \naf C_1$}
\quad
and

\qquad
\mbox{$A_2;x \!\la\! B_2, \naf C_2$}
\quad
infer

\qquad\qquad
$A_1;A_2 \la B_1,B_2, \naf C_1, \naf C_2$

\item[S-HYP]
From
\mbox{$A_1 \la B_1, \naf x_1, \naf C_1$,}

\qquad\qquad\qquad
\dots,

\quad\qquad\qquad
\mbox{$A_n \la B_n, \naf x_n, \naf C_n,$}

\quad\qquad\qquad
\mbox{$A \la x_1,\dots,x_n, \naf C$ \quad infer}

\qquad
\mbox{$A_1;\dots;A_n \la$}

\qquad\qquad
\mbox{$B_1,\dots,B_n, \naf C_1,\dots,\naf C_n,\naf A, \naf C$}
\end{description}
Several of these rules are analogous to or similar to well-known rules in
the literature.
For example, {\bf Nonmin} is weakening;
{\bf WGPPE} is analogous to cut; and
{\bf S-HYP} is a version of hyper-resolution.
Let $\vdashs$ denote the consequence relation generated by these
rules, for convenience allowing sets of rules on the right hand side of
$\vdashs$.
Then $P \lrarrows P'$ abbreviates $P \vdash_s P'$ and $P' \vdash_s P$.
As well, define
\[
\Cns{\alphabet}{P}
=
\{
r \in {\cal L}_\alphabet \mid P \vdashs r\}.
\]
%
%Last, for program $P$ and rule $r$ we have:
Then the above set of inference rules is sound and complete with respect to the entailment $\models_s$.
\begin{theorem}[\cite{Wong2008}]
$P \incls r$ iff $P \vdashs r$.
\end{theorem}

\section{The Approach}
\label{sec:approach}
\subsection{Formal Preliminaries}
\label{sec:preliminaries}

Since forgetting in our approach amounts to decreasing the alphabet,
or signature, of a logic program, we need additional notation for relating
signatures.
Let $\alphabet$ and $\alphabet'$ be two signatures where
$\alphabet' \subset \alphabet$.
Then $\alphabet'$ is a \emph{reduction}%
\footnote{The standard term in model theory is \emph{reduct}
\cite{ChangKeisler2012,Doets96,Hodges97}.
However \emph{reduct} has its own meaning in ASP, and so we adopt this
variation.}
of $\alphabet$, and $\alphabet$ is an \emph{expansion} of $\alphabet'$.
%Furthermore, if $w \in \SE{\alphabet}$ and $w' \in \SE{\alphabet'}$ where
Furthermore, if $w$ is an SE interpretation on $\alphabet$ and $w'$ is an SE interpretation on $\alphabet'$ where
$w$ and $w'$ agree on the interpretation of symbols in $\alphabet'$ then
$w'$ is the \emph{$\alphabet$-reduction} of $w$, and $w$ is an
\emph{$\alphabet'$-expansion} of $w'$.
For fixed $\alphabet' \subset \alphabet$, reductions are clearly unique
whereas expansions are not.

For a logic program $P$, $\Sig{P}$ denotes the signature of $P$, that is,
the set of atoms mentioned in $P$.
SE models are defined with respect to an understood alphabet;
for SE model $w$ we also use $\Sig{w}$ to refer to this alphabet.
Thus for example if $\alphabet = \{a,b,c\}$ then, with respect to
$\alphabet$, the SE model $w = (\{a\},\{a,b\})$ is more perspicuously written
as $(\{a, \neg b, \neg c\},\{a,b, \neg c\})$, and so in this case
$\Sig{w} = \{a,b,c\}$.

If $\alphabet' \subset \alphabet$ and for SE models $w$, $w'$ we have
$\Sig{w} = \alphabet$ and $\Sig{w'} = \alphabet'$ then we use
$\reduce{w}{\alphabet'}$ to denote the reduction of $w$ with respect to
$\alphabet'$ and we use $\expand{w'}{\alphabet}$ to denote the set of
expansions of $w'$ with respect to $\alphabet$.
This notation extends to sets of models in the obvious way.
As well, we use the notion of a reduction for logic programs;
that is, for $\alphabet' \subseteq \alphabet$, 
\[
\reduce{P}{\alphabet'} =
\{ r \in P \mid \Sig{r} \subseteq \alphabet' \}.
\]

\subsection{An Abstract Characterisation of Forgetting}

As described, our goal is to define forgetting with respect to the logical
content of a logic program.
For example, if we were to forget $b$ from the program
$\{ a \la b., \; b \la c.\}$, we would expect the rule $a \la c$ to be in the
result, since it is implicit in the original program.
Consequently, our primary definition is the following.
\begin{definition}
\label{def:forget}
Let $P$ be a disjunctive logic program over signature $\alphabet$.  
The result of \emph{forgetting $\alphabet'$ in $P$},
denoted $\F(P,\alphabet')$, is given by:
\[
\F(P,\alphabet') = \Cns{\alphabet}{P} \cap {\cal L}_{\alphabet \setminus \alphabet'}.
\]
\end{definition}
That is, the result of forgetting a set of atoms $\alphabet'$ in program $P$
is simply the set of SE consequences that of $P$ over the original
alphabet, but excluding atoms from $\alphabet'$.

This definition is very simple. 
This characterization is abstract, at the \emph{knowledge level}.
As a consequence, many formal results are very easy to show.
On the other hand, the definition is not immediately practically useful
since forgetting results in an infinite set of rules.  
Consequently a key question is to determine a finite characterisation (that
is to say, a uniform interpolant) of $\F$.
We explore these issues next.

The following results are elementary, but show that the definition of
forgetting has the ``right'' properties.
\begin{proposition}\ 
\label{prop:basic}
Let $P$ and $P'$ be disjunctive logic program and let $\alphabet$ (possibly
primed or subscripted) be alphabets.

\begin{enumerate}
\item
$P \vdashs \F(P, \alphabet)$

\item
If $P \lrarrows P'$ then
$\F(P, \alphabet) \lrarrows \F(P', \alphabet)$

\item
\(
\F(P, \alphabet)
=
\Cns{\alphabet'}{\F(P, \alphabet)}
\)

\qquad
where $\alphabet' = \Sig{P}\setminus\alphabet$.

\item
\(
\F(P, \alphabet)
=
\)

\qquad
\(
\F(\F(P, \alphabet \setminus \{a\}), \{a\}))
\)

\item
\(
\F(P, \alphabet_1 \cup \alphabet_2)
=
\)

\qquad
\(
\F(\F(P, \alphabet_1), \alphabet_2))
\)

\item
$P$ is a \emph{conservative extension} of $\F(P, \alphabet)$.  
\end{enumerate}
\end{proposition}
Thus, forgetting results in no consequences not in the original theory.
As well, the result of forgetting is independent of syntax and yields
a deductively-closed theory (Parts 2 and 3).
Part 4 gives an iterative means of determining forgetting on an
element-by-element basis.
The next part, which generalises the previous, shows that forgetting is
decomposable with respect to a signature, which in turn implies that
forgetting is a commutative operation with respect to its second argument.
Last, $P$ is a conservative extension of the result of forgetting, which is
to say, trivially $\Sig{P} \setminus \alphabet' \subseteq \Sig{P}$, and the
consequences of $P$ and $\F(P, \alphabet)$ coincide over the language
${\cal L}_{\Sig{P}\setminus \alphabet'}$.

With regards to SE models, we obtain the following results giving an
alternative characterisation of forgetting.
Here only we use the notation $\SEs{\alphabet}{P}$ to indicate the SE
models of program $P$ over alphabet $\alphabet$.
\begin{proposition}\ 
\label{prop:forgetModels}
%  JD: Prior to this point we used $\alphabet \subseteq \alphabet'$. We
%  should be consistent in the usage...
Let $\alphabet' \subseteq \alphabet$, and let
$\Sig{P} \subseteq \alphabet$.
\begin{enumerate}
\item
\(
\SEs{\alphabet \setminus \alphabet'}{\F(P, \alphabet')}
=
\reduce{\SEs{\alphabet}{P}}{(\alphabet \setminus \alphabet')}
\)

\item
\(
\SEs{\alphabet}{\F(P, \alphabet')}
=
\expand{(\reduce{\SEs{\alphabet}{P}}{(\alphabet \setminus \alphabet')})}{\alphabet}
\)
\end{enumerate}
\end{proposition}

\noindent
The first part provides a semantic characterisation of forgetting:
the SE models of $\F(P, \alphabet')$ are exactly the SE models of $P$
restricted to the signature $\alphabet \setminus \alphabet'$.
Very informally, what this means is that the SE models of
$\F(P, \alphabet')$ can be determined by simply dropping the symbols in
$\alphabet'$ from the SE models of $P$.
The second part, which is a simple corollary of the first, expresses
forgetting with respect to the original signature.

Of course, one may wish to re-express the effect of forgetting in the
original language of $P$; 
in fact, many approaches to forgetting assume that the underlying language is
unchanged.
To this end, we can consider a variant of Definition~\ref{def:forget} as
follows, where $\alphabet' \subseteq \alphabet$.
\begin{equation}
\label{eqn:AltDefn}
\F_\alphabet(P, \alphabet') \equiv \Cns{\alphabet}{\F(P,\alphabet')}
%\quad
%\mbox{ where $\alphabet = \Sig{P}$}.
\end{equation}
That is, $\F(P,\alphabet')$ is re-expressed in the original language with
signature $\alphabet$.
The result is a theory over the original language, but where the resulting
theory carries no contingent information about the domain of application
regarding elements of $\alphabet'$.
%\footnote{There is a sublety here.
%Consider forgetting the proposition $p$, but re-expressing the result in
%the original language including $p$.
%In propositional logic, there would indeed be no contingent information
%concerning $p$.
%In an epistemic logic, there could be subjective knowledge such as
%$K \neg K p$, but no contingent knowledge of the domain of application.}

The following definition is useful in stating results concerning
forgetting.
\begin{definition}
\label{def:irrelevant}
Signature $\alphabet$ is \emph{irrelevant} to $P$, $\IR(P, \alphabet)$, iff
there is $P'$ such that $P \lrarrows P'$ and
$\Sig{P'} \cap \alphabet = \emptyset$.%
\end{definition}

Zhang and Zhou \shortcite{ZhangZhou2009} give four postulates
characterising their approach to forgetting in the modal logic S5.
An analogous result follows here with respect to forgetting re-expressed in
the original signature:
\begin{proposition}\ 
\label{prop:postulates}
Let $\alphabet' \subseteq \alphabet$ and let
$\Sig{P}$, $\Sig{P'} \subseteq \alphabet$.

Then $P' = \F_\alphabet(P, \alphabet')$ iff
\begin{enumerate}
\item
$P \vdashs P'$

\item
If $\IR(r, \alphabet')$ and $P \vdashs r$ then $P' \vdashs r$

\item
If $\IR(r, \alphabet')$ and $P \not\vdashs r$ then
$P' \not\vdashs r$

\item
$\IR(P', \alphabet')$
\end{enumerate}
\end{proposition}

\noindent
For the last three parts we have that, if a rule $r$ is independent of
a signature $\alphabet'$, then forgetting $\alphabet'$ has no effect on whether
that formula is a consequence of the original knowledge base or not
(Parts 2 and 3).
The last part is a ``success'' postulate:
the result of forgetting $\alphabet'$ yields a theory expressible without
$\alphabet'$.  

\subsection{A Finite Characterisation of Forgetting}
\subsubsection{Aside: Forgetting in Propositional Logic}

We first take a quick detour to forgetting in propositional logic to
illustrate the general approach to finitely characterising forgetting.  
Let $\phi$ be a formula in propositional logic and let $p$ be an atom;
the standard definition for forgetting $p$ from $\phi$ in propositional logic
is defined to be $\phi[p/\top] \vee \phi[p/\bot]$.
It is not difficult to show that this is equivalent to
Definition~\ref{def:forget}, but suitably re-expressed in terms of
propositional logic.
This definition however is not particularly convenient.
It is applicable only to finite sets of formulas.
As well, it results in a formula whose main connective is a disjunction.

An alternative is given as follows.
Assume that a formula (or formulas) for forgetting is expressed in clause
form, where a (disjunctive) clause is expressed as a set of literals.
For forgetting an atom $p$, consider the set of all clauses obtained by
resolving on $p$:
\begin{definition}
Let $S$ be a set of propositional clauses and $p \in {\cal P}$.
Define
\begin{eqnarray*}
Res(S,p) =&&
\hspace{-.5cm}
\{ \phi \mid
\exists \phi_1, \phi_2 \in S \mbox{ such that } \;
\\
&&
p \in \phi_1 \mbox{ and } \neg p \in \phi_2, \; \mbox{ and }
\\
&&
\phi =
(\phi_1 \setminus \{p\}) \cup (\phi_2 \setminus \{\neg p\})
\}
\end{eqnarray*}
\end{definition}

\noindent
We obtain the following, where $Forget_{P\!C}$ refers to forgetting in
propositional logic:
%:
\begin{theorem}
\label{thm:forgetPC}
Let $S$ be a set of propositional clauses over signature ${\cal P}$ and
$p \in {\cal P}$.
\[
\F_{P\!C}(P,p)
\;\lrarrow\;
\reduce{{S}}{({\cal P}\setminus \{p\})} \cup Res(S,p).
\]
\end{theorem}

This provides an arguably more convenient means of computing forgetting,
in that it is easily implementable, and one remains with a set of clauses.

\subsubsection{Back to Forgetting in Logic Programming:} 
We can use the same overall strategy for computing forgetting in a
disjunctive logic program.
In particular, for forgetting an atom $a$, we can use the inference rules
from \cite{Wong2008} to compute ``resolvents'' of rules that don't mention
$a$.
It proves to be the case that the corresponding definition is a bit more
intricate, since it involves various combinations of {\bf WGPPE} and
{\bf S-HYP}, but overall the strategy is the same as for propositional logic.

In the definition below, $ResL\!P$ corresponds to $Res$ for forgetting in
propositional logic.
In propositional logic, $Res$ was used to compute all resolvents on an
atom $a$.
Here the same thing is done: we consider instances of {\bf WGPPE} and
{\bf S-HYP} in place of propositional resolution; 
these instances are given by the two parts of the union, respectively,
below.

\begin{definition}
Let $P$ be a disjunctive logic program and $a \in \alphabet$.

Define:
\begin{eqnarray*}
\lefteqn{ResL\!P(P,a) = }
\\
&&
\{ r \mid
\exists r_1, r_2 \in P \mbox{ such that } \;
\\
&&
\quad
r_1 = A_1 \!\la\! B_1, a, \naf C_1,
\\
&&
\quad
r_2 = A_2;a \!\la\! B_2, \naf C_2,
\\
&&
\quad
r = A_1;A_2 \la B_1,B_2, \naf C_1, \naf C_2
\;\}
\\
&&
\cup
\\
&&
\{ r \mid
\exists r_1,\dots,r_n, r' \in P \mbox{ such that } a = a_1 \;
\\
&&
\quad
r_i = A_i \!\la\! B_i, \naf a_i, \naf C_i,
\quad
1 \leq i \leq n
\\
&&
\quad
r' = A \la a_1, \dots a_n, \naf C
\quad
\mbox{and}
\\
&&
\quad
r = 
A_1;\dots;A_n \la
\\
&&
\qquad\qquad
B_1,\dots,B_n, \naf C_1,\dots,\naf C_n,\naf A, \naf C
\;\}
\end{eqnarray*}
\end{definition}
\noindent
We obtain the following:
\begin{theorem}
\label{thm:forgetLP}
Let $P$ be a disjunctive logic program over $\alphabet$ and $a \in \alphabet$.
Assume that any rule $r \in P$ is satisfiable, non-tautologous, and contains
no redundant occurrences of any atom.

Then:

\(
\F(P,a)
\lrarrows
\reduce{{P}}{(\alphabet\setminus \{a\})} \;\cup\; ResL\!P(P,a).
\)
\end{theorem}
\noindent
{\bf Proof Outline:}
From Definition~\ref{def:forget}, $\F(P,a)$ is defined to be the set of
those SE consequences of program $P$ that do not mention $a$.
Thus for disjunctive rule $r$, $r \in \F(P,a)$ means that $P \vdashs r$ and
$a \not\in \Sig{r}$.
Thus the left-to-right direction is immediate:
Any $r \in \reduce{{P}}{(\alphabet\setminus \{a\})}$ or $r \in ResL\!P(P,a)$ 
is a SE consequence of $P$ that does not mention $a$.

For the other direction, assume that we have a proof of $r$ from $P$,
represented as a sequence of rules.
If no rule in the proof mentions $a$, then we are done.
Otherwise, since $r$ does not mention $a$, there is a last rule in the
proof, call it $r_n$ that does not mention $a$, but is obtained from rules
that do mention $a$.
The case where $r_n$ is obtained via {\bf Taut}, {\bf Contra}, or
{\bf Nonmin} is easily handled.
If $r_n$ is obtained via {\bf WGPPE} or {\bf S-HYP} then there are rules
$r_k$ and $r_l$ that mention $a$ (and perhaps other rules in the case of
{\bf S-HYP}).
If $r_k$,$r_l \in P$ then $r_n \in ResLP(P,a)$.
If one of $r_k$, $r_l$ is not in $P$ (say, $r_k$) then there are several cases,
but in each case it can be shown that the proof can be transformed to
another proof where the index of $r_k$ in the proof sequence is decreased
and the index of no rule mentioning $a$ is increased.  
This process must terminate (since a proof is a finite sequence), where the
premisses of the proof are either rules of $P$ that do not mention $a$,
elements of $ResL\!P(P,a)$, or tautologies.  

Consider the following case, where
$r_n = A_1;A_2;A_3 \la B_1,B_2,B_3$, and we use the notation that each
$A_i$ is a set of implicitly-disjoined atoms while each $B_i$ is a set of
implicitly-conjoined literals.
Assume that $r_n$ is obtained by an application of {\bf WGPPE} from 
$r_k = a;A_1;A_2 \la B_1,B_2$ and $r_l = A_3 \la a,B_3$.
Assume further that $r_k$ is obtained from
$r_i = a;b;A_1 \la B_1$ and $r_j = A_2 \la b,B_2$
by an application of {\bf WGPPE}.
This situation is illustrated in Figure~1a.

\begin{center}
\includegraphics[width=0.40\textwidth]{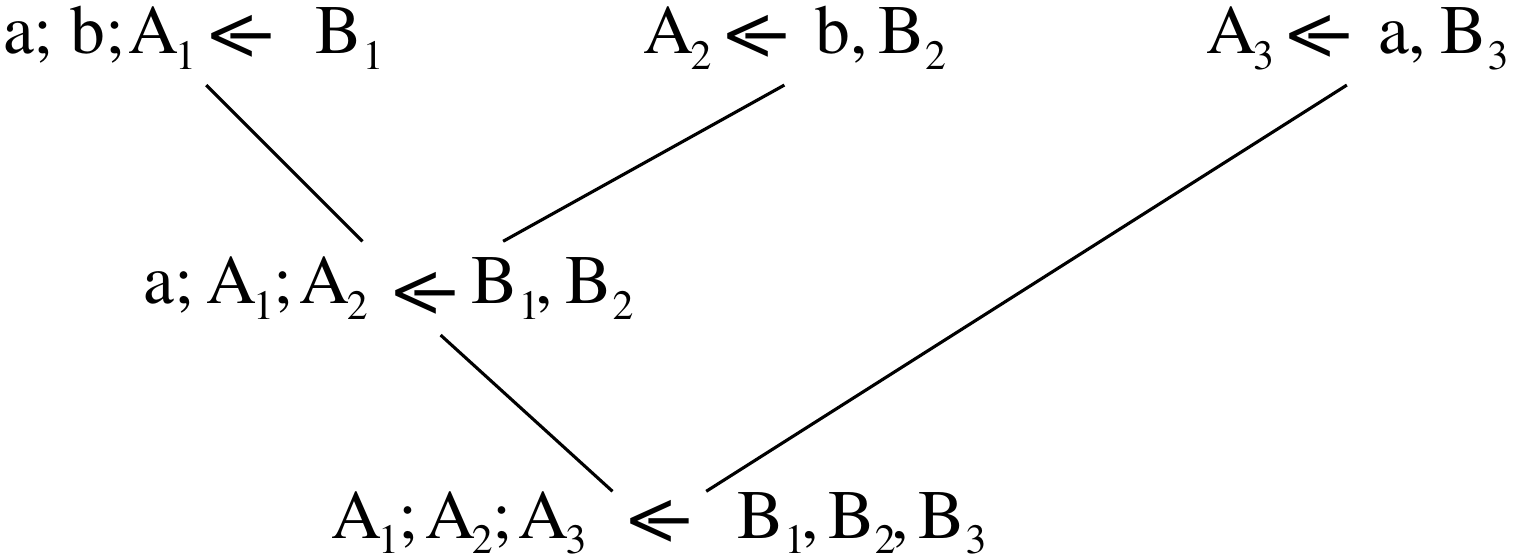}

\smallskip
Figure 1a
\end{center}
\medskip

Then essentially the steps involving the two applications of {\bf WGPPE}
can be ``swapped'', as illustrated in Figure~1b, where $r_k$ is replaced by
$r_k' = a;A_1;A_2 \la B_1,B_2$.

\medskip
\begin{center}
\includegraphics[width=0.40\textwidth]{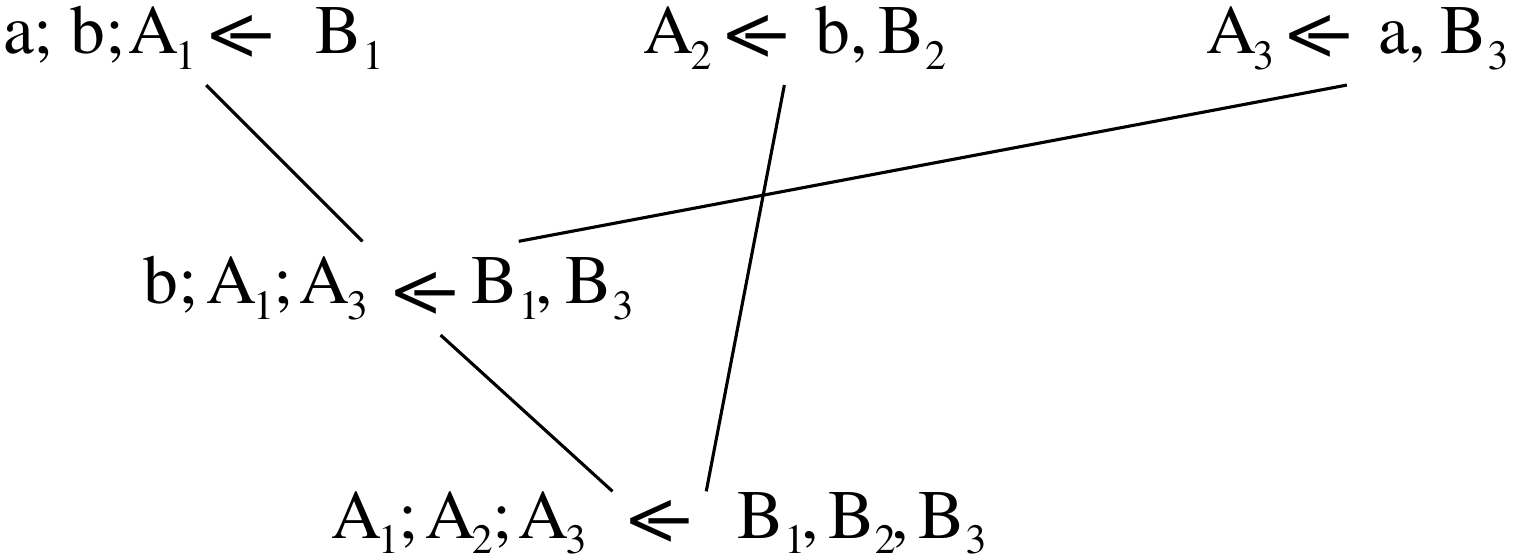}

\smallskip
Figure 1b
\end{center}

\medskip
Thus the step involving $a$ is informally ``moved up'' in the proof.
There are 12 other cases, involving various combinations of the inference
rules, but all proceed the same as in the above.  
$\Box$

\medskip

The theorem is expressed in terms of forgetting a single atom.
Via Proposition~\ref{prop:basic}.4 this readily extends to forgetting a set
of atoms.
Moreover, since we inherit the results of Propositions~\ref{prop:basic}
and~\ref{prop:postulates}, we get that the results of forgetting are
independent of syntax, even though the expression on the right hand side
of Theorem~\ref{thm:forgetLP} is a set of rules obtained by transforming
and selecting rules in $P$.
It can also be observed that forgetting an atom results in at worst a
quadratic blowup in the size of the program.  
While this may seem comparatively modest, it implies that forgetting a set
of atoms may result in an exponential blowup.

\begin{example}
Let $P = \{p \la \naf q.\;\; r \la p\}$.
Forgetting $p$ yields $\{ r \la \naf q\}$ (where $r \la \naf q$ is obtained
by an application of {\bf WGPPE}), while forgetting $q$ and $r$ yield
programs $\{ r \la p \}$ and $\{ p \la \naf q \}$ respectively.
\end{example}

\subsection{Computation of Forgetting}

By Theorem~\ref{thm:forgetLP}, we have the following algorithm for computing the result of forgetting.
A rule $r$ is a tautology if it is of the form $r = A; b \la b, B, \naf C$; 
a rule $r$ is a contradictory if it is of the form $r = A; c \la B, \naf c, \naf C$; 
a rule $r$ is minimal if there is no rule $r'$ in $P$ such that $B(r')\subseteq B(r)$, $H(r')\subseteq H(r)$ and one of these two subset relations is proper; otherwise, $r$ is non-minimal.

\vskip 1mm

\noindent {\bf Algorithm 1 (Computing a result of forgetting)}

\noindent {\em Input}: Disjunctive program $P$ and literal $a$ in
$P$.

\noindent {\em Output}: $\F(P, a)$.

\noindent {\em Procedure:}

{\em Step 1.} Remove tautology rules, contradiction rules and non-minimal rules from $P$. The resulting disjunctive program is still denoted $P$.

{\em Step 2.} Collect all rules in $P$ that do not contain the atom  $a$, denoted $P'$.

{\em Step 3.}  For each pair of rules 
$r_1 = A_1 \!\la\! B_1, a, \naf C_1$
and
$r_2 = A_2;a \!\la\! B_2, \naf C_2$,
add the rule
$r = A_1;A_2 \la B_1,B_2, \naf C_1, \naf C_2$ to $P'$

{\em Step 4.} For each rule $r' = A \la a_1, \dots a_n, \naf C$ where for
some $i$, $a_i=a$, and for each set of $n$ rules
$\{r_i = A_i \!\la\! B_i, \naf a_i, \naf C_i \;|\; 1 \leq i \leq n\}$,
add the rule
$r = A_1;\dots;A_n \la B_1,\dots,B_n, \naf C_1,\dots,\naf C_n,\naf A, \naf C$
to $P'$.

{\em Step 5.} Return $P'$ as $\F(P, a)$.

%For each pair of rules $r'$ and $r_1$ with $a\in B^+(r')$ and $a\in B^-(r_1)$,
%add the rule $H(r_1)\lrarrows B(r_1)-(\naf a), B(r')-a,\naf H(r')$.

\vskip 1mm
%Here, for a set of literals $L$ and a literal $l$, $L-l$ is obtained from $L$ by removing $l$. 
Some remarks for the algorithm are in order. Obviously, Step 1 is to preprocesss the input program by eliminating tautology rules, contradiction rules and non-minimal rules from $P$. Initially, all rules that do not contain $a$, which are trivial SE-consequences of $P$, are included in the result of forgetting.
% This step is simple but key to the efficiency of the computation procedure since forgetting is needed only to apply on the part of input program in which each rule contains the atom $a$. 
In many practical applications, such a part of input program is usually not very large and thus forgetting can be efficiently done although the input program can be very large.
Step 3 and Step 4 implement two resolution rules {\bf WGPPE} and {\bf S-HYP}, respectively. 
%However, Step 4 is a special version of {\bf S-HYP}. For this reason, the correctness of this algorithm is not straightforward. 

%
%%%%%%%%%%%%
%\begin{theorem}\label{thm:alg:improved}
%For any disjunctive program $P$ and an atom $a$, Algorithm 1 outputs
%$\F(P,a)$.
%\end{theorem}
%%%%%%%%%%%%%%%
%

%\begin{proof}
%???
%\end{proof}

\subsection{Conflict Resolving by Forgetting: Revisited}

\cite{EiterWang06,EiterWang08} explore how their semantic forgetting for logic programs can be used to resolve conflicts in multi-agent systems.
However, their notion of forgetting is based on answer sets and thus does
not preserve the syntactic structure of original logic programs, as pointed out in \cite{ChengERSW06}.
In this subsection, we demonstrate how this shortcoming of Eiter and Wang's forgetting can be overcome in our SE-forgetting for disjunctive programs.
    
The basic idea of conflict resolving \cite{EiterWang06,EiterWang08} consists of two observations:

\begin{enumerate}
\item
each answer set corresponds to an agreement among some agents;

\item
conflicts are resolved by forgetting some literals/concepts for
some agents/ontologies.
\end{enumerate}
%
%%%%%%%%%%%%%%%%%
\begin{definition}
Let $\calS = (P_1, P_2,\ldots, P_n)$, where each logic program $P_i$
represents the preferences/constraints of Agent $i$. A {\em
compromise} of $\calS$ is a sequence $C=(F_1,F_2,\ldots,F_n)$ where
each $F_i$ is a set of atoms to be forgotten from $P_i$. An {\em
agreement} of $\calS$ on $C$ is an answer set of
$\forget{\calS}{C}=\forget{P_1}{F_1}\cup \forget{P_2}{F_2}\cup
\cdots\cup \forget{P_n}{F_n}$.
\end{definition}
%%%%%%%%%%%%%%%%%
%
For specific applications, we may need to impose certain conditions on each $F_i$.
However, the two algorithms (Algorithms 1 and 2) in \cite{ChengERSW06} may
not produce intuitive results if directly used in a practical application.
Consider a simple scenario with two agents.
%
%%%%%%%%%%
\begin{example}\cite{ChengERSW06} \label{exa:pref:simple}
Suppose that two agents A1 and A2 try to reach an agreement on
submitting a paper to a conference, as a {\em regular paper} or as a {\em
system description}. If a paper is prepared as a system description,
then the system may be implemented either in Java or Prolog. The
preferences and constraints are as follows.
\begin{enumerate}
\item The same paper cannot be submitted as both a regular paper and system description.
\item A1 would like to submit the paper as a regular one and,
in case the paper is submitted as a system description and there is
no conflict, he would prefer to use Java.
\item A2 would like to submit the paper as a system description but not
prefer regular paper.
% due to some reasons.
\end{enumerate}
Obviously, the preferences of these two agents are jointly
inconsistent and thus it is impossible to satisfy both at the same
time. The scenario can be encoded as a collection of three
disjunctive programs ($P_0$ stands for general constraints): \(
\calS = (P_0, P_1, P_2) \) where $R, S, J, P$ mean ``regular
paper,'' ``system description,'' ``Java'' and ``Prolog,''
respectively: $P_0 =\{\la R,S\},
    %%%%%%%%%%%%%%%%%%
P_1 =\{R \la \dru J \la S, \naf P\},
    %%%%%%%%%%%%%%%%%%
P_2 =\{\la R \dru S \la\}$.

Intuitively, if $A_1$ can make a compromise by forgetting $R$, then
there will be an agreement $\{S, J\}$, that is, a system description
is prepared and Java is used for implementing the system. However,
if we directly use forgetting in conflict resolution, by forgetting
$R$, we can only obtain an agreement $\{S\}$ which does not contain
$J$. In fact, this is caused by the removal of $J\la S,\naf P$ in
the process of forgetting. This rule is abundant in $P_1$ but
becomes relevant when we consider the interaction of A1 with other
agents (here A2).
\end{example}

As pointed out in \cite{ChengERSW06}, it is necessary to develop a theory of forgetting for disjunctive programs such that {\em locally abundant}
(or {\em locally irrelevant}) rules in the process of forgetting can be preserved.
Our SE forgetting provides an ideal solution to the above problem.
This can be seen from the definition of SE-forgetting and Algorithm 1 (if needed, we don't have to eliminate non-minimal rules in Step 1).
In fact, $\F(P_1, R) = \{J \la S, \naf P\}$, which preserves the locally redundant rule $J \la S, \naf P$.

%%%%%%
%\subsection{{\sf SE-Forget}: A prototype implementation}
%{\sf SE-Forget} is implemented in Java and 

%\section{Discussion}
%\label{sec:discussion}
%
%Prospects for forgetting in other forms of logic programs?
%Normal LPS would be nice, as would general LPs.

\section{Conclusion}

In this paper we have addressed forgetting under SE models in disjunctive logic programs,
wherein forgetting amounts to a reduction in the signature of a program. 
Essentially, the result of forgetting an atom (or set of atoms) from a
program is the set of SE consequences of the program that do not mention
that atom or set of atoms.
This definition then is at the \emph{knowledge level}, that is, it is 
abstract and is independent of how a program is represented.
Hence this theory of forgetting is useful for tasks such as knowledge base
comparison and reuse. A result of the proposed forgetting under SE models is also a result of forgetting under answer sets but not vice versa.
Moreover, we have developed an efficient algorithm for computing forgetting
in disjunctive logic programs, which is complete and sound with respect to
the original knowledge-level definition.

A prototype implementation, of forgetting has been implemented in Java and
is available publicly at \url{http://www.ict.griffith.edu.au/~kewen/SE-Forget/}.
While our experiments on the efficiency of the system are still underway,
preliminary results show that the algorithm is very efficient.
Currently we are still working on improving efficiency of the implementation
and are experimenting on applying it to large practical logic programs and
randomly generated programs.
We plan to apply this notion of forgetting to knowledge base comparison and
reuse.
For future work we also plan to investigate a similar approach to
forgetting for other classes of logic programs.

\bibliography{ai}
\bibliographystyle{named}

\end{document}